\documentclass[11pt,a4paper]{article}
\usepackage[hyperref]{acl2019}
\usepackage{times}
\usepackage{latexsym}
\usepackage{graphicx}
\usepackage{xr}
\usepackage{url}
\usepackage{hyperref}

\aclfinalcopy 

\title{Argument Invention from First Principles}

\author{
Yonatan Bilu , %
Ariel Gera, %
Daniel Hershcovich, Benjamin Sznajder, Dan Lahav, \\
\textbf{
Guy Moshkowich, Anael Malet, Assaf Gavron, Noam Slonim %
}  \\
IBM Research\\
\{yonatanb,arielge,benjams,dan.lahav,guy.moshkowitch,assaf.gavron,noams\}@il.ibm.com,\\
daniel.hershcovich@gmail.com,
anael.malet@sciencespo.fr
}

\date{}

\begin{document}
\maketitle
\begin{abstract}
  Competitive debaters often find themselves facing a challenging task -- how to debate a topic they know very little about, with only minutes to prepare, and without access to books or the Internet? What they often do is rely on ''first principles'', commonplace arguments which are relevant to many topics, and which they have refined in past debates.
  In this work we aim to explicitly define a taxonomy of such principled recurring arguments, and, given a controversial topic, to automatically identify which of these arguments are relevant to the topic. 
  As far as we know, this is the first time that this approach to argument invention is formalized and made explicit in the context of NLP.
  The main goal of this work is to show that it is possible to define such a taxonomy. While the taxonomy suggested here should be thought of as a ''first attempt'' it is nonetheless coherent, covers well the relevant topics and coincides with what professional debaters actually argue in their speeches, and facilitates automatic argument invention for new topics.
\end{abstract}

\section{Introduction}

In his treatise {\em De Inventione} Cicero defines the five canons of classical rhetoric as: inventio (invention),  dispositio (arrangement), elocutio (style), memoria (memory), and pronuntiatio (delivery). The first of these, {\em Inventio}, is defined as a systematic search for arguments \cite{st-martin}, with applicability to a wide variety of situations often seen as a desired property \cite{invention}. This problem has been referred to in the context of NLP as the task of {\em Argument Invention} \cite{walton2012, walton2017}, but did not receive abundant attention.

One natural way people go through the process of {\em inventio} is to look for arguments in relevant texts, or, if they are familiar with the topic, rely on their knowledge and {\em memoria} for doing so. This is reminiscent of the way Argument Mining algorithms operate (see e.g. \citealp{lippi2015argument, lippi2016argumentation}). However, we often find ourselves in situations where that is not possible. For example, when arguing politics over lunch, we might find ourselves backed into a corner, facing a topic with which we are not very familiar, but somehow nonetheless need to justify or oppose. This often happens because we were initially arguing some {\em principle}, and now we need to apply it to an unfamiliar example.

Professional debaters often face this problem. 
Presented with an unfamiliar topic they need to quickly come up 
with relevant arguments. The main technique for doing so is called arguing from ''first principles'' -- relying on a ''bank'' of principled arguments, which are relevant to a wide variety of topics\footnote{This is reminiscent of, yet different from, the Aristotelian use of this term, which refers to self-evident propositions.}. 

A common example is the ”Black market” argument:
banning a product or a service may lead
to the creation of a black market, which in
turn makes products or services obtained therein
less safe, leads to exploitation, attracts criminal elements,
and so on. Hence, even if we 
agree that something should not be encouraged, it is advisable to have it legal and regulated. 

This kind of argument can be made, {\em mutatis mutandis}, when debating quite different topics, such as legalizing organ trade, or banning pornography. 
However, it is not always relevant when debating whether to legalize or ban something; For example, when debating whether to legalize polygamy or ban breastfeeding in public, the black market argument seems less appropriate.


Here, we aim to create a knowledge base of such principled arguments, which, when given a topic for debate or a critical essay, would readily yield the relevant ones. We do this in a framework of certain types of {\em motions} (section \ref{sec:def}). Specifically, we define several commonplace themes which are likely to be a point of contention -- that is, where arguments of opposing stance can be made around this theme\footnote{In the context of debates, these are called ''clashes''.}. We show that for most motions there exist relevant arguments within the suggested knowledge base, and that they can be identified automatically with reasonable precision and recall. Moreover, we show that professional human debaters often allude to such arguments when they debate.

\section{Related work}
Previous computational work on argument invention was mainly done within the field of argument mining, where -- as the name implies -- the focus is on identifying arguments within a given text. Most works (e.g. \citealp{stab2014annotating, palau2009argumentation, eger2017neural}) assume that a relevant text is provided, while some include the task of extracting such text from a large open-domain corpus (e.g. \citealp{levy2014coling, rinott2015show, shnarch2018will, al2016cross, levy2018towards}). The work here complements such techniques by providing a  dataset of arguments whose manual construction facilitates automatic retrieval for topics of interest and ensures quality, validity, style and so on.

A somewhat similar approach is suggested in work of \citet{walton2017, walton2018}, where arguments are constructed from a database \cite{reed2004araucaria} of smaller argumentative building blocks. However, these building blocks are topic-specific and can not readily provide arguments for topics not in the database.

The attempt to categorize arguments by looking for commonalities dates back to ancient times, such as Aristotle's list of 28 topoi \cite{rhetoric}. Modern works, such as \citet{perelman1971new, walton2008argumentation, walton2013argumentation}, expanded on these ideas, similarly focusing on how an argument's conclusion is inferred from its premises. Unlike these efforts, the taxonomy suggested here is of recurring principled semantic themes. That is, arguments which in this work would be categorized as belonging to a specific argument theme could be of various topoi and follow different argumentation schemes.

In modern competitive debating the notion of commonalities between topics is prevalent due to the advantages they serve in overcoming knowledge barriers and in speeding up argument generation\footnote{E.g. \url{https://debate.uvm.edu/dcpdf/WUDC\%20Malaysia_2014_Debating_and_Judging_Manual.pdf}}. Armed with limited facts on the topic, the task of locating recurring patterns in order to argue the motion abstractly is composed of understanding what are the fundamental 'clashes' in the debate (cf. \citet{sonnreich2012monash}, ''debating from first principles''), similar to the taxonomy herein.

Our approach bears similarities to work in social sciences that attempts to describe different types of information {\em framing}, usually in the context of the news media (e.g. \citealp{semetko2000framing}). Recurring themes, like {\em Fairness and equality} or {\em Crime and punishment}, can be identified in the way the news media frames a certain policy issue or event \cite{card2015framingcorpus}. 
\citet{devreese2005framingtypology} differentiates between specific and generic frames, characterizing the latter as those that can be applied to a wide range of events and contexts. Similarly, our work aims to categorize commonplace themes and identify their relevance (at a considerably larger scale), in the context of framing a topic that is subject to debate. 

Our work also has some commonalities with psychological research on ideology
(e.g. \citealp{altemeyer1981right,sidanius2001social,jost2003fair}). For example, \citet{everett201312} lists a 12-item scale to assess conservative ideology - of these, some map to our taxonomy (e.g. Welfare), while others are too specific. Moreover, conservatism in itself gives rise to one class of recurring arguments in our work.

\section{Definitions}\label{sec:def}
In the context of parliamentary debate, a {\em motion} is a proposal that is to be deliberated by two sides (government and opposition). Here we formally define a motion as a pair {\em (action, topic)}, where {\em topic} is a Wikipedia title (or a redirect to one), and {\em action} is a term coming from a closed set of allowed actions (Appendix \ref{app:actions}), and describes the government's proposal w.r.t. {\em topic}. For example, the motion {\em (ban, smoking)} should be interpreted as the government suggesting to ban smoking, and can be explicitly phrased as ''we should ban smoking''\footnote{In the context of competitive debate, the phrasing would usually be ''This House would ban smoking''.}. Note that not all combinations of {\em action} and {\em topic} make for a good motion -- the implied proposal should be worth deliberating; one for which reasonable 
arguments can be made by either side.

One often discerns between {\em policy motions} where the government proposes a concrete policy, and {\em analysis motions} in which the government declares its opinion on the {\em topic}. For the sake of simplicity and brevity (and with some abuse of notation) we 
will ignore this distinction. In particular, one of the allowed ''actions'' is {\em brings more harm than good}, which would usually be considered as indicating an analysis motion. 

We define a {\em Class of Principled Arguments}, or {\em CoPA}, as a set of arguments revolving around a principled recurring theme (we define the name of the CoPA as this theme), alongside a set of motions to which this theme is relevant. Formally, a CoPA is a pair {\em c = (A, M)}, where $A$ is a set of arguments, and $M$ is a set of motions, s.t. every $a \in A$ is an argument that can plausibly be made when deliberating any motion $m \in M$. For every $a \in A$ we say that $a$ is an argument in $c$, and similarly that every $m \in M$ is a motion in $c$, and that $m$ and $c$ {\em match}. 

In this work we focus on modelling debate {\em clashes}, and hence are interested in $A$'s which contain arguments of opposing stances towards the class's theme. For the sake of simplicity we consider only $A$'s of size $2$, and only simple arguments, which are essentially just claims or premises (appendix \ref{app:classes}). Note that this is indeed a simplification, and that for many CoPAs several distinct arguments can be naturally included in $A$ (cf. Section \ref{sec:discussion}). 

The pair of claims may directly contradict each other, denoting a disagreement about facts. But by and large we tried to select pairs of claims that people tend to agree with\footnote{This is somewhat similar to what \citet{perelman1971new} call ''Values'' and also close to what Aristotle calls {\em Endoxa} \cite{rhetoric}.}, but would assign different valuation depending on their point of view (see \citealp{kock2009choice}). For example, one of the CoPAs we define is {\em Clean energy}, with 
$A$ = \{''Humanity must embrace clean energy in order to fight climate change'', ''Ecological concerns add further strain on the economy''\}, and $M$ including motions such as {\em (subsidize, renewable energy)} and {\em (fight, global warming)}. Most people would agree that on the one hand climate change is a problem, and on the other that moving towards clean energy will be expensive. When debating motions where clean energy is a relevant theme the two sides are likely to agree that both claims have some merit, yet disagree on which supersedes the other. The valuation might very well depend on their subjective viewpoints, but also on the specific motion.

\section{Data}\label{sec:data}
\subsection{Initial data}
The definition above is a functional one, oriented towards facilitating labeling motion-class matches, so some care is required in CoPA construction. A pair of claims with one saying that the policy will not work, and the other that it will, defines a CoPA that essentially covers all policy motions. Conversely, very particular claims will yield 
a class for which relevant motions are hard to come by. 

The set discussed in this work was defined based on the following guidelines:
\begin{enumerate}
    \item One is able to define two concise claims of opposing stance towards the CoPA's theme.\label{req:clash}
    \item One can think of at least three motions (not necessarily from the initial set) which would belong to the CoPA, and are not overly similar to one another.\label{req:recurring}
\end{enumerate}
Our motivation for requirement \ref{req:clash} was to model ''clashes'', the recurring themes in debates which are points of contention, since our main use-case is, given a motion, to suggest argumentative text with a clear stance towards the motion. Other use cases may relax this requirement, according to their goals. Requirement \ref{req:recurring} ensures that the CoPA indeed captures a recurring theme, rather than a specific one. 

Two annotators were presented with these guidelines and an initial list of 100 motions to make the task more concrete\footnote{see supplementary materials in \url{www.research.ibm.com/haifa/dept/vst/debating_data.shtml}}. They authored a list of about 60 CoPAs which was manually curated by two of the authors to a more concise, final list of 37 CoPAs, 
to avoid redundancy and ease the 
following 
labeling task.

Appendix \ref{app:classes} lists these 37 CoPAs and the claims 
therein. They are quite varied -- some revolve around public policy (e.g. {\em Environment, Public health}), others on basic rights and freedoms (e.g. {\em Right to privacy, Freedom of religion}), some on the effect of a policy (e.g. {\em Black market, Greater good}), while yet others are very general ({\em Fixable, Conservatism}, and {\em Framework})\footnote{When analyzing our data, it will sometimes be interesting to omit these three classes, to ascertain they do not significantly skew the results.}.

Next, the same two annotators annotated all 100 motions for membership in the suggested CoPAs. 
In total, 92 motions were matched to at least one CoPA, and on average each motion was matched to 2.03 CoPAs. In our dataset, the greatest number of CoPAs a motion is a member of is 5, with two motions achieving this number -- (legalize, prostitution) and (ban, infant circumcision). The full annotation is provided in the supplementary materials.

In order to validate this annotation, a sample of motion-CoPA pairs was annotated via crowd-sourcing platform Figure-Eight\footnote{https://www.figure-eight.com/}. For each motion, argument pairs from 2 randomly chosen non-matching CoPAs and (up to) 2 randomly chosen matching ones were annotated by 5 labelers. Average inter-annotator Cohen's kappa agreement was 0.63. Then, taking the majority vote for each pair as the crowd-sourced label, we computed agreement between it and our initial labeling, yielding a kappa score of 0.78. These indicate a rather high agreement, especially in the context of computational argumentation \cite{passonneau2014benefits,habernal2016argument}.

\subsection{Expanded data}
The initial construction of CoPAs was done with the aim of identifying themes which are recurrent in general, not just in the initial 100 motions. To verify that they generalize to other motions, we collected 589 additional motions, 
and annotated them for CoPA membership (the same annotators who did the initial annotation). On this new dataset, we found that 503 motions were matched to at least one CoPA (85\%), and on average each motion was matched to 1.94 CoPAs. Hence, while our modeling may be biased by the initial set of motions, it 
seems to generalize 
well to new ones.

As with the initial set of motions, we used crowd-sourced annotations of a similarly-sampled portion of the dataset to verify the full annotation, 
attaining an average inter-annotator kappa score of 0.60, and a kappa score of 0.76 when comparing the majority vote to the full annotation. 

The full dataset can be found in the supplementary material.

\subsection{CoPA claims in recorded speeches}\label{sec:recorded-speeches}
It is natural to ask whether the claims authored for each CoPA are an artificial construct for facilitating motion assignment, or are actual claims, likely to 
be made by 
people deliberating these motions. To this end we considered the speeches we recorded in \citet{listening-comprehension}. Each such speech is given in the context of a motion, all of which are included in our dataset. For each motion we extracted the CoPAs to which it belongs according to our annotation, yielding 184 speeches with at least one matching CoPA. 7 Figure-Eight annotators were presented with speeches in both audio and written form, alongside the claims 
from the matching CoPAs. They were asked whether each claim was (i) explicitly made by the speaker, was (ii) implicit in the speech or was (iii) not mentioned at all. A total of 800 (speech, claim) 
pairs were annotated, 
with one half of them being claims of a stance opposing that of the speaker.

In order to analyze agreement between annotators, we considered (i) and (ii) as a positive label and (iii) as negative. The average inter-annotator Cohen Kappa score was 0.54. Moreover, since we showed both CoPA claims to the annotators we checked whether claims whose stance opposed that of the speaker were ever marked positive. With only 5\% of the annotations being so, we concluded that the annotation was of reasonable quality (cf. section \ref{sec:res-rec}).

\section{Matching Methods}\label{sec:matching-methods}
Having a sizable dataset of (motion, CoPA) pairs, we examined several classifiers over it. That is, given a motion and a CoPA, the classifier aims to determine whether they match. Since the CoPAs are quite varied, we examined various classifiers, some focused on a motion's action, some on its topic, and some on a combination of both. 

Specifically, we examined the following classifiers:\\
    \textbf{By action (BA-$k$):} Some actions are strongly indicative for (some of) the CoPAs a motion belongs to. To utilize this, this classifier trains by computing, for each allowed action $a$, and each CoPA $c$, the probability $p(c,a)$ that a motion with action $a$ will belong to $c$. Prediction for a new motion $m = (a,t)$ is done by assigning each CoPA $c$ the score $p(c,a)$. In addition, if the number of (training-set) motions in $c$ with action $a$ is less than some parameter $k$, this method makes no prediction.\\  
    \textbf{By topic, nearest neighbors (KNN):} Given a left-out motion, $m = (a,t)$ the algorithm goes over the motions $m_i = (a_i, t_i)$ in the training set, looking for those such that $t_i$ is most similar to $t$ (using the similarity measure of \citealp{term-relater}). It keeps only those whose similarity is above a threshold of $0.5$. If there are less than 3 such motions, no prediction is made. Otherwise it takes the (at most) top 5 motions. For each CoPA $c$, the assigned score is the fraction of these  motions which belongs to $c$. This is then used to predict membership. \\
    \textbf{By topic, word2vec features (W2V):} Each motion $m = (a,t)$ is represented as the word2vec \cite{word2vec} embedding vector of $t$ (if $t$ is a multi-word expression the vectors are summed and normalized). This vector is then used as a feature vector for a logistic regression classifier. That is, each CoPA is assigned the classification score of the classifier so trained. As a safeguard mechanism, we also determine an actions blacklist, $B_c$, for each CoPA $c$. An action $a$ is in $B_c$ if in the training set no motion with action $a$ is in $c$. During prediction, if the left-out motion's action is in $B_c$, it will not be predicted as belonging to $c$.\\
    \textbf{By topic, Naive Bayes (NB) and Recurrent Neural Network (RNN):} Following the work of \citet{abstractness} each motion $m = (a,t)$ is associated with a set of retrieved sentences containing the term $t$. For a given CoPA $c$, all the sentences associated with motions in $c$ are considered as positive examples, and those of motions not in $c$ as negative examples. For {\em NB}, A Naive Bayes classifier is then trained over the unigrams of these sentences, and uses its score for prediction. In addition, the same blacklist safeguard mechanism as for {\em W2V} above is used.\\
    Similarly, these sentences were used to train an RNN to differentiate between positively-labeled sentences and negatively-labeled ones. See \citet{abstractness} for more details on these methods.\\ 
    \textbf{By topic and action (LR):} We defined 17 features based on similarities between a motion and a CoPA, and on co-occurrence counts, similar to the one used in {\em BA-k}. A logistic regression classifier was trained and scored on the resulting feature vectors over pairs of (motion, CoPA). See appendix \ref{app:features} for details.\\
    \textbf{Ensemble:} For completeness, all 6 methods above were aggregated by simply assigning each CoPA the highest score it attained among all of them. 
    We note that this is a very naive approach; while all methods produce scores in $[0, 1]$, it is not clear that they are comparable. In practice one would probably use an aggregation method that differentiates between the different classification methods, and between different CoPAs. 

All classifiers (except one\footnote{For technical reasons we trained and evaluated the RNN method using 3-fold cross-validation.}) were evaluated in a leave-one-motion-out framework, over all motions and over relevant CoPAs. That is, each classifier was trained and tested 689 times -- in each iteration it was trained over 688 motions and the relevant CoPAs, and then predicted whether the left-out motion matched these CoPAs. More precisely, each CoPA is assigned a score. We vary the score threshold, and determine membership by whether the assigned score exceeds the threshold.

\section{Results}
\subsection{Complete dataset}
\begin{figure}[ht!]
\centering
\includegraphics[width=75mm]{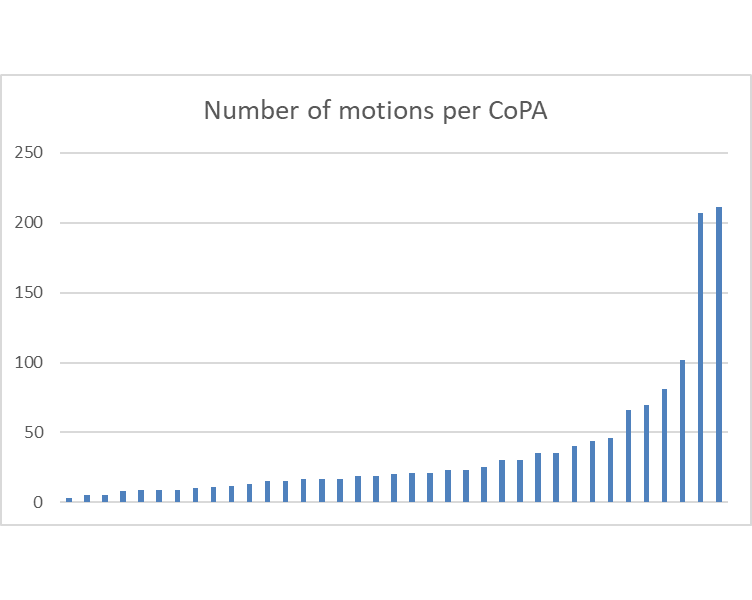}
\caption{Distribution of the number of motions per CoPA. \label{fig:hist}}
\end{figure}

All in all, our dataset describes the motion-CoPA relations of 689 motions and 37 CoPAs. Figure \ref{fig:hist} shows a histogram of the CoPA sizes in this dataset. The two biggest CoPAs ({\em Fixable} and {\em Conservatism}) include nearly one third of the motions (207 motions and 211 motions respectively), while at the other end, the class {\em Self determination} contains only 3 motions. Most CoPAs (32 out of the 37) are of modest size, containing less than 10\% of the motions. 

\begin{figure*}[ht!]
\centering
\includegraphics[width=160mm]{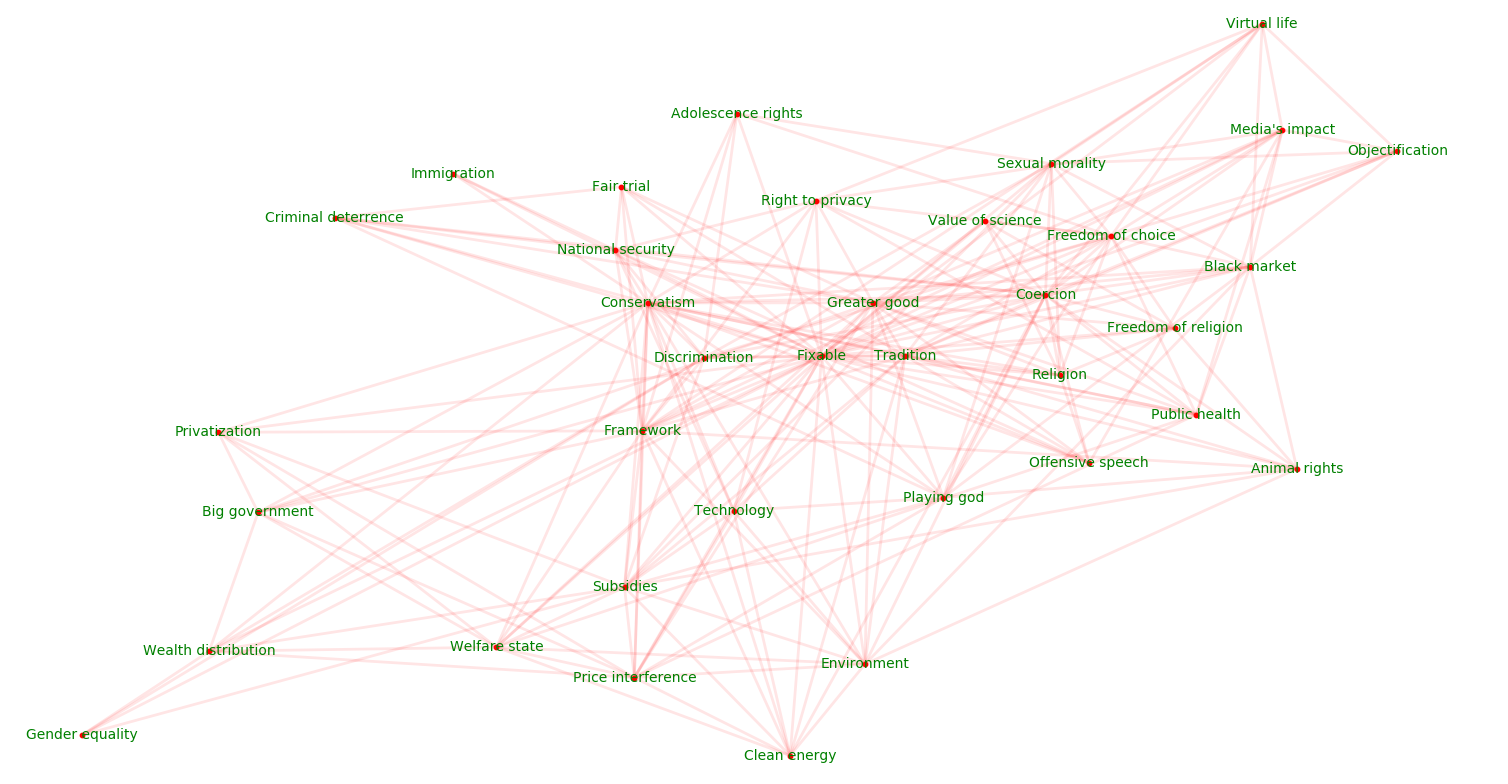}
\caption{Graph of CoPAs, where edges indicate non-empty intersection and distance between vertices is indicative of intersection size. Not shown is ''Self determination'', an isolated vertex. \label{fig:copa-graph}}
\end{figure*}

\begin{figure}[ht!]
\centering
\includegraphics[width=75mm]{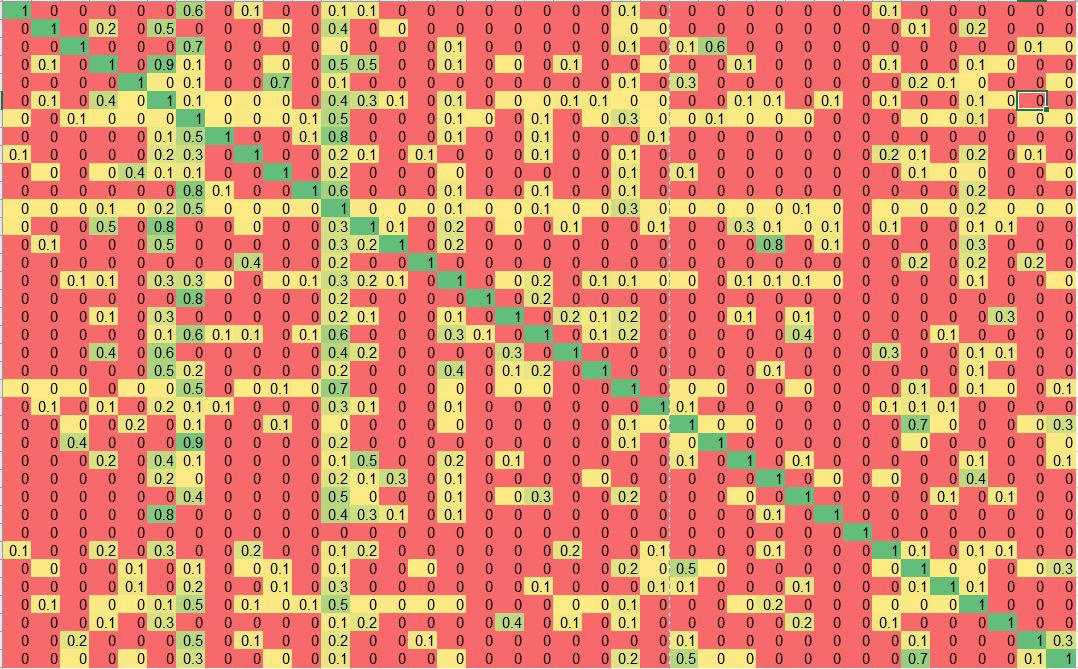}
\caption{Fraction of overlapping motions among classes; the value of entry (i,j) is the fraction of motions in class $i$ which also appear in class $j$. Green indicates high values, red low ones. \label{fig:overlap}}
\end{figure}

Importantly, the CoPAs capture different facets of a motion, rather than induce a partition of the motions set. On average, a CoPA has a non empty intersection with 11.95 other CoPAs, with the average intersection size being 21\% of the CoPA size. Figure \ref{fig:copa-graph} shows the inter-connectivity graph among CoPAs. The aforementioned CoPA {\em Self determination} is an isolated vertex in this graph, but other than that the graph is connected. This is especially noteworthy, considering that many of the CoPAs are rather small. Figure \ref{fig:overlap} shows a heat-map of overlap sizes.

In the complete dataset, 87\% of the motions belong to at least one CoPA, and on average each motion belongs to 1.95 CoPAs. That is, while this is certainly only a first step toward modeling principled recurring arguments, the suggested CoPAs are indeed a concise set that covers distinct argumentative themes and offers a good coverage w.r.t. the world of motions defined here.

\subsection{CoPA claims in recorded speeches}\label{sec:res-rec}
Of the 184 annotated speeches of \citet{listening-comprehension}, 87\% had at least one CoPA-claim annotated as positive\footnote{A pair is considered positive if a majority of annotators chose option (i) or (ii); cf. section \ref{sec:recorded-speeches}.}, and in total, 66\% of the 400 (speech, claim) pairs (where the stance of the claim and the speaker were aligned) were marked as positive. However, in the vast majority of cases the claim was marked as implicit in the speech -- according to the annotation only 10\% of the speeches contain a CoPA-claim explicitly, and only 5\% of the pairs are labeled as an explicit mention.

One reason for this may be the three ''general'' CoPAs, since their claims are so general that they would usually be at least implicit in a speech. When removing these CoPAs from the analysis 62\% of the speeches have at least one positive claim, and 39\% of the pairs are positive. Hence, even without these classes, most speeches implicitly mention at least one claim from the dataset. This is probably due to the rather generic phrasing of the claims, which in the first place were constructed to be applicable ''as-is'' in multiple contexts. In other words, this annotation not only confirms that the CoPA claims convey arguments actually alluded to by humans, but that they do so at a rather high level, and so capture arguments that are not only plausible for a motion but also probable.

Conversely, for each CoPA, we also examined the speeches to which it matched, and computed the fraction of these speeches in which the CoPA's claim was annotated as positive.
 Of the 37 CoPAs, 29 match motions in \citet{listening-comprehension}. For all but one ({\em Sexual morality}), in at least 25\% of the relevant speeches, the CoPA's claim (of the correct stance) was labeled positive. For 24 CoPAs at least 50\% were so labeled.

\subsection{Motion-CoPA matching}\label{subsec:match-res}
As noted in section \ref{sec:matching-methods}, we evaluated the proposed matching methods in a leave-one-out framework. For the action-based method, {\em BA-k}, we set $k=5$ and consider only CoPAs which contains at least $5$ motions with the same action. For the topic-based methods we considered only CoPAs which were manually marked as topic-based (see appendix \ref{app:classes}) and contain at least 10 motions. The {\em LR} method was naturally evaluated on all CoPAs. 

\begin{figure}[ht!]
\centering
\includegraphics[width=75mm]{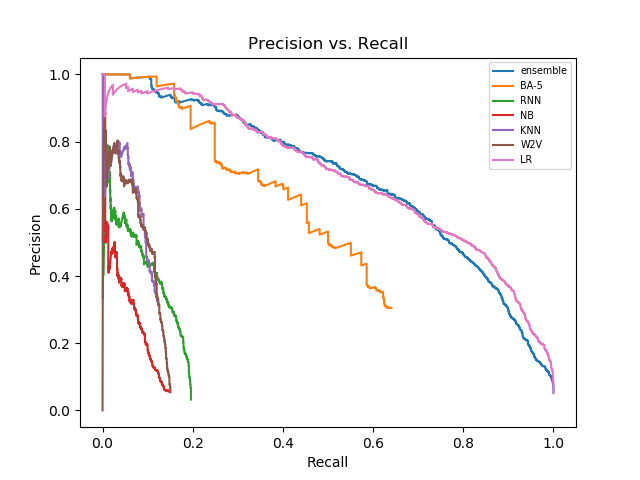}
\caption{Precision-Recall curve for the various motion-CoPA matching methods. \label{fig:prec-recall}}
\end{figure}

Figure \ref{fig:prec-recall} describes the precision-recall trade-off for each of the 7 methods from section \ref{sec:matching-methods}, which is computed over all (motion, CoPA) pairs: precision is the fraction of matching pairs whose score is above the threshold from among all pairs with such a score; recall is the ratio between the number of matching pairs with such a score, and the total number of matching pairs. 

Note that for methods which look at only a subset of the CoPAs recall is bound to be low, since recall calculation takes into consideration all CoPAs, not just this subset.

\begin{figure}[ht!]
\centering
\includegraphics[width=75mm]{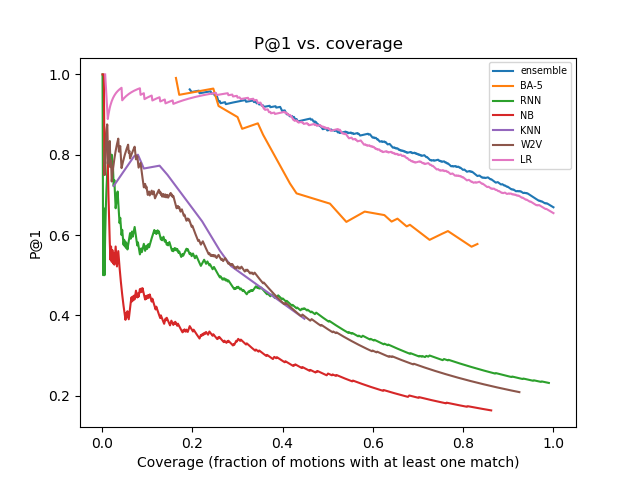}
\caption{P@1 vs. coverage of the various motion-CoPA matching methods. \label{fig:pAt1}}
\end{figure}

With the task of {\em Argument Invention} in mind, a use-case of interest is, given a motion, to provide (at least) one CoPA from which argumentative content can be extracted. Accordingly, Figure \ref{fig:pAt1} evaluates the precision for the highest scoring CoPA of each motion -- for a given threshold, the figure depicts the fraction of motions whose highest scoring CoPA is both a match and above the threshold, as a function of the fraction of motions for which at least one CoPA passes the threshold. As can be seen, for a threshold that yields CoPA prediction for half the motions, the ensemble method has $86\%$ precision for its top prediction.

\begin{figure}[ht!]
\centering
\includegraphics[width=75mm]{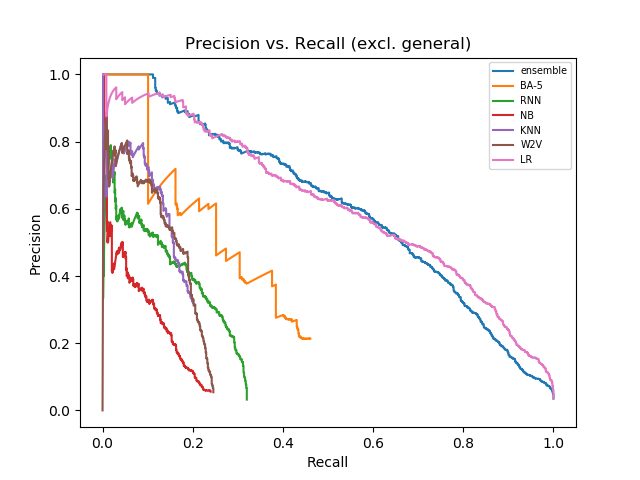}
\includegraphics[width=75mm]{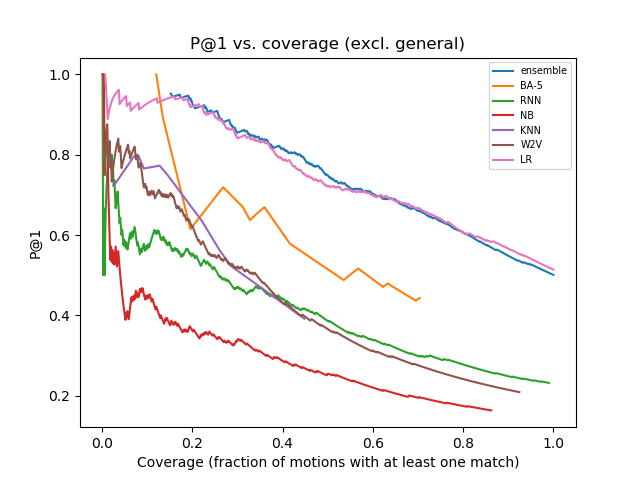}
\caption{Precision results when ignoring the three general classes, {\em Conservatism, Fixable, Framework} \label{fig:no-general}}
\end{figure}

Finally, recall that the three ''general'' CoPAs ({\em Conservatism, Fixable, Framework}) might dominate the predictions analyzed above. Omitting them from the analysis does reduce precision somewhat, but nonetheless, the top prediction of the ensemble method for a threshold yielding a prediction for half the motions attains a precision of $75\%$ (Figure \ref{fig:no-general}; For this analysis the ''general'' classes were not included in the recall computation).

A naive baseline would always (and only) predict the CoPA with the largest number of motions as a match. When considering all CoPAs, this attains a precision of $30\%$ (for {\em Conservatism}), and when omitting the three general CoPAs, a precision of $12\%$ (for {\em Coercion}).

\section{Discussion}\label{sec:discussion}
The most basic argument model is probably Aristotle's {\em categorical syllogism}, which consists of a major premise, a minor premise and a conclusion \cite{rhetoric}; with the minor premise being a categorical proposition connecting between the major premise and the conclusion. The canonical example is: 
\begin{quote}
    All men are mortal.\\
    Socrates is a man.\\
    Therefore, Socrates is mortal.
\end{quote}
It is interesting to consider the CoPAs and the claims they contain in this context. When aiming to argue for a motion, and identify a CoPA to which it belongs, one can create a syllogistic-like argument as follows. The major premise would be the CoPA claim, the minor premise would explain why the motion is a member of the CoPA, and the conclusion would be that the motion should stand\footnote{Since the major premise here is not a categorical proposition, the argument will not 
be 
true in the propositional, {\em modus ponens} sense. But if the claim is indeed an {\em endoxa}, the argument should be one that most people consider plausible.}.

For example, when deliberating the motion (further exploit, solar energy), and identifying that it belongs to the CoPA {\em Clean energy}, the resulting (heuristic) argument could be:
\begin{quote}
    Humanity must embrace clean energy in order to fight climate change.\\
    Solar energy is a form of clean energy.\\
    Therefore, humanity must further exploit solar energy.
\end{quote}

Similarly, a very basic model for describing deliberation is Hegelian dialectics:
The deliberation or debate starts with a thesis, which is countered by an antithesis, and is then resolved with synthesis. The CoPA's claims can be seen as providing a thesis and an antithesis in the context of a member motion, with the synthesis dependent on the motion and on the valuation of the claims by the adjudicator.

A major challenge in constructing the CoPAs herein was finding an explicit phrasing for the arguments, one that would be suitable without further context. One example for this is the {\em backlash} argument -- an argument stating that implementing the policy will be counter-productive, since it will create a backlash reaction. While this is a common argument, arguing why a backlash reaction will occur and how it will be counterproductive may well depend on context. Moreover, it is difficult to phrase an appropriate claim of the opposite stance without further context.  

However, the CoPAs can actually provide us with an appropriate context needed for phrasing such arguments. Thus, one could phrase more specific backlash arguments for the CoPA {\em Subsidies} or {\em Coercion} (with different phrasings), and use them when the CoPA is matched to a motion. In other words, we can expand the set of CoPA arguments to include more than just 2 claims; it could include further instances of principled arguments, each perhaps tailored to the specific CoPA. Recall that our motivation is {\em Argument Invention} -- when a CoPA is matched, the underlying system can present all arguments the CoPA contains.

Indeed, with the aim of assisting critical writing in mind, one need not limit CoPAs to claims or even to coherent arguments. CoPAs could very well be rhetorical {\em loci} for relevant anecdotes, proverbs, memes, quotes from famous people and so on. They could also include text written in different styles to accommodate different types of presentations ({\em pronuntiatio}). In response to a topic, a system making use of the CoPAs knowledge base could present all these texts to the users, or filter them according to their preferences.

An interesting research direction in this respect is to  include in a CoPA, for each claim it contains, a rebuttal argument that counters it. This can enable an argumentative dialog system \cite{rach2018utilizing}, along the lines alluded to in \citet{listening-comprehension} -- one can envision a system that performs {\em listening comprehension}, identifies the relevant CoPAs, checks whether any of the claims in the CoPA were mentioned in the audio, and, for those that do, responds with the rebuttal arguments matching this claim. This is similar to scripted dialog systems, with the important difference that the texts are not written for a specific scenario. They are principled arguments which can be used in many different contexts, allowing for an open-domain dialog system. We intend to describe such a system in future work.

Furthermore, a CoPA can include complex argumentative structures such as those in AraucariaDB \cite{reed2004araucaria}, from which multi-layered arguments can be constructed, e.g. using the Carneades Argumentation System of \citet{walton2012}. That is, instead of having such data per topic, as is currently the case in AraucariaDB, having such data for commonplace principled arguments facilitates their use over a wide range of topics. Note that for this the stance of the argument w.r.t the CoPA and the motion is important. For the sake of simplicity and brevity we have ignored this issue in this manuscript, but the relevant stance labeling is available in the supplementary material.

In the field of computational argumentation, {\em de novo} argument synthesis has received relatively little attention. One naive attempt is that of \citet{claim-generation}, where claims are generated by pasting together a topic and short predicate. 
The framework suggested here may provide a richer and more stable basis for argumentative text generation. That is, a CoPA may include structured data which describes its principal theme. Then, when presented with a motion in this CoPA, the system would automatically generate, {\em de novo}, argumentative text based on this structured data \emph{and the topic}. For example, this could be an NLG neural net trained on a large corpus of claims extracted using argument mining for motions in the CoPA. 

Finally, let us reappraise the basic intuition of the corpus-wide argument-mining approach to argument invention -- that an effective argumentation is one that draws on the widest possible array of proofs and arguments. Rhetoricians have characterized the art of convincing as starting from general and basic views, facts and opinions accepted by everyone \cite{perelman1971new,kock2009choice}. 
In other words, an efficient argument starts not from the most original and unseen premises, but from what the audience takes as consensual, and only then progresses to what is controversial.
Therefore, the need for principled arguments is not only a question of time and practicality, but also stems from the essential nature of rhetoric: it is the necessity to call on the general views and opinions shared by everyone and to show that they uphold the desired conclusion.

\section{Conclusion}
We presented a novel framework -- 689 controversial motions with a variety of topics and actions -- in which the {\em Argument Invention} task can be formalized
and 
assessed. We formalized the notion of commonplace principled arguments, and 
suggested 
a concrete and diverse taxonomy for them. While this taxonomy can certainly be expanded and refined, 
it nonetheless has the basic desired properties: most motions in our framework belong to it, annotators tend to agree on CoPA-motion matching, this matching can be done automatically with reasonable success, and human debaters tend to allude to the ascribed arguments when debating these motions.

\section{Acknowledgments}
We thank Uri Zakai for insightful discussions which helped initiate this work. We thank Anatoly Polnarov and Noga Zaslawsky for their help in developing the taxonomy and data sets presented here. 

\bibliography{main}
\bibliographystyle{acl_natbib}

\clearpage



\newpage
\appendix
\section{Allowed actions}
\label{app:actions}
\begin{table}[ht!]
\begin{tabular}{ |l|c|  }
\hline
\small Action & \small \#motions \\
\hline
\small a criminal offence & \small 3 \\
\small abandon & \small 52 \\
\small abolish & \small 57 \\
\small adopt & \small 45 \\
\small ban & \small 90 \\
\small brings/bring/brought more good than harm & \small 3\\
\small brings/bring/brought more harm than good & \small 3\\
\small cancel & \small 13 \\
\small close & \small 4 \\
\small criminalize & \small 4 \\
\small decrease & \small 2 \\
\small disband & \small 29 \\
\small discourage & \small 6 \\
\small encourage & \small 6 \\
\small end & \small 33 \\
\small exaggerated & \small 2 \\
\small fight & \small 25 \\
\small fight for & \small 5 \\
\small further exploit & \small 21 \\
\small increase & \small 49 \\
\small introduce & \small 22 \\
\small justified & \small 6 \\
\small legalize & \small 16 \\
\small limit & \small 23 \\
\small lower & \small 5 \\
\small mandatory & \small 18 \\
\small nationalize & \small 6 \\
\small not abandon & \small 2 \\
\small not ban & \small 2 \\
\small not mandatory & \small 1 \\
\small not subsidize & \small 6 \\
\small not tax & \small 2 \\
\small oppose & \small 3 \\
\small privatize & \small 13 \\
\small prohibit & \small 17 \\
\small protect & \small 19 \\
\small raise & \small 4 \\
\small reduce & \small 2 \\
\small subsidize & \small 58 \\
\small support & \small 6 \\
\small tax & \small 3 \\
\small unjustified & \small 3 \\
\hline
\end{tabular}
\end{table}

\section{Classes of Principled Arguments}\label{app:classes}
This appendix lists the 37 CoPAs. In each, the number of motions associated with it follows the class's name, and two opposing CoPA-claims are listed in the entries below it. CoPAs marked with a superscript $t$ are those deemed as topic-related.

\begin{tabular}{ |p{3cm}|p{3cm}|  }
 \hline
 \multicolumn{2}{|c|}{Adolescent rights \small (9 motions)} \\
 \hline
 \small Many adolescents can not make responsible decisions &
 \small Adolescents are as capable as adults\\
 \hline
 \multicolumn{2}{|c|}{Animal rights$^t$ \small (21 motions)} \\
 \hline
 \small Animals should not be treated as property &
 \small There is nothing wrong with using animals to further human interests\\
 \hline
 \multicolumn{2}{|c|}{Big government \small (21 motions)} \\
 \hline
 \small Public utility is best served by actions coordinated by central government &
 \small Public interest is best served and propelled by voluntary interactions, and not ones dictated by government\\
 \hline
 \multicolumn{2}{|c|}{Black market$^t$ \small (35 motions)} \\
 \hline
 \small Prohibiting products and activities makes them less visible and available, and thus less harmful & 
 \small Prohibition is counterproductive and only leads to increased demand\\
 \hline
 \multicolumn{2}{|c|}{Clean energy$^t$ \small (25 motions)} \\
 \hline
 \small Humanity must embrace clean energy in order to fight climate change & 
 \small Ecological concerns add further strain on the economy\\
 \hline
 \multicolumn{2}{|c|}{Coercion \small (81 motions)} \\
 \hline
 \small A decisive and enforced policy is the best way to deliver a message &
 \small Enforcement tends to be less effective than persuasion and education\\
 \hline
 \multicolumn{2}{|c|}{Conservatism \small (211 motions)} \\
 \hline
 \small The current system is working, and making such a change could have negative consequences &
 \small It is time to change the old ways and try something new\\
 \hline
 \multicolumn{2}{|c|}{Criminal deterrence \small (8 motions)} \\
 \hline
 \small When people will have to pay for their actions there will be less crime &
 \small Strict punishment is not effective in preventing criminal behavior\\
 \hline
\multicolumn{2}{|c|}{Discrimination$^t$ \small (19 motions)} \\
 \hline
 \small It is a fact that there are differences between people. Hence, there should sometimes be differences in the way people are treated. &
 \small All people should be treated equally\\
 \hline
 \multicolumn{2}{|c|}{Environment$^t$ \small (44 motions)} \\
 \hline
 \small People must protect nature and respect its biological communities &
 \small Environmentalism stands in the way of technological progress and economic growth\\
 \hline
 \end{tabular}

\begin{tabular}{ |p{3cm}|p{3cm}|  }
 \hline
\multicolumn{2}{|c|}{Fair trial$^t$ \small (17 motions)} \\
 \hline
 \small Upholding the rights of the accused and ensuring a fair process is the best way to maintain justice &
 \small The focus should not be on the rights of criminals, but on protecting the law-abiding public from harm\\
 \hline
\multicolumn{2}{|c|}{Fixable \small (207 motions)} \\
 \hline
 \small There are some issues here that need addressing, but that doesn't mean it should all be eliminated &
 \small Due to the many problems associated with [TOPIC], the best course of action would be to put an end to it\\
 \hline

 \multicolumn{2}{|c|}{Framework \small (102 motions)} \\
 \hline
 \small [TOPIC] works efficiently &
 \small [TOPIC] fails to achieve its goals \\
 \hline
 \multicolumn{2}{|c|}{Freedom of choice \small (35 motions)} \\
 \hline
 \small People have the right to make their own choices, including bad ones &
 \small It is the duty of society to protect people from their own bad choices\\
 \hline
 \multicolumn{2}{|c|}{Freedom of religion \small (10 motions)} \\
 \hline
 \small People should be free to practice their religion &
 \small Religions are outdated, irrational and harmful\\
 \hline
\multicolumn{2}{|c|}{Gender equality$^t$ \small (5 motions)} \\
 \hline
 \small Banishing the misguided notions of gender roles in society is the way to achieve true equality for women &
 \small Gender roles reflect true biological differences between the sexes. It doesn't make sense to ignore them\\
 \hline
 \multicolumn{2}{|c|}{Greater good \small (40 motions)} \\
 \hline
 \small The safety and well-being of the community is more important than individual freedom &
 \small Individual freedom is a sacred value. It cannot be subordinated to subjective opinions deciding what is best for society\\
 \hline
 \multicolumn{2}{|c|}{Immigration$^t$ \small (5 motions)} \\
 \hline
 \small People who come in search of a safer and better life should not be turned away &
 \small Mass immigration threatens social cohesion\\
 \hline
 \multicolumn{2}{|c|}{Media's impact$^t$ \small (15 motions)} \\
 \hline
 \small Media consumption has no significant social or behavioral effects &
 \small Stereotypes distributed in the media lead to a distorted view of society and of the other\\
 \hline
 \multicolumn{2}{|c|}{National security \small (19 motions)} \\
 \hline
 \small Some rights and freedoms need to be limited in the interest of national security &
 \small Security does not justify brushing aside fundamental rights and freedoms\\
 \hline
 \multicolumn{2}{|c|}{Objectification \small (9 motions)} \\
 \hline
 \small Women should have the power to use and show their bodies as they would like to &
 \small Society cannot allow women to be treated as commodities\\
 \hline
  \end{tabular}

\begin{tabular}{ |p{3cm}|p{3cm}|  }
 \hline
\multicolumn{2}{|c|}{Offensive speech \small (13 motions)} \\
 \hline
 \small Freedom of expression is meaningless if it does not apply to troubling and controversial ideas &
 \small The freedom of expression does not legitimize offending people's values and beliefs\\
 \hline
\multicolumn{2}{|c|}{Playing god \small (20 motions)} \\
 \hline
 \small People can, and therefore should, interfere with nature in order to take care of their needs &
 \small Only God should determine how life comes into being and how it comes to an end\\
 \hline
 \multicolumn{2}{|c|}{Price interference \small (46 motions)} \\
 \hline
 \small Price regulation is useful for achieving social and economic goals &
 \small Market forces should determine the rates of prices and fees\\
 \hline
 \multicolumn{2}{|c|}{Privatization \small (30 motions)} \\
 \hline
 \small Privatization often leads to improved efficiency and quality &
 \small The state is a better and a more natural provider of public goods and services\\
 \hline
 \multicolumn{2}{|c|}{Public health$^t$ \small (17 motions)} \\
 \hline
 \small It is the government's duty to safeguard public health and promote healthy life choices &
 \small The state should have no role in encouraging or discouraging particular lifestyle choices\\
 \hline
 \multicolumn{2}{|c|}{Religion$^t$ \small (23 motions)} \\
 \hline
 \small Religion creates a sense of community for people, and organizes human life &
 \small Religion has proven over the years that it is a harmful, destructive force\\
 \hline
 \multicolumn{2}{|c|}{Right to privacy$^t$ \small (23 motions)} \\
 \hline
 \small The right to privacy is a fundamental right &
 \small Privacy is not absolute. There are instances when it must be compromised in order to protect society\\
 \hline
 \multicolumn{2}{|c|}{Self-determination \small (3 motions)} \\
 \hline
 \small The political status of a territorial entity should be defined by its population &
 \small Self-determination cannot be handed freely, especially not as a prize for violence\\
 \hline
 \multicolumn{2}{|c|}{Sexual morality \small (17 motions)} \\
 \hline
 \small Sexual morality must be protected by opposing immoral lifestyles, and not ignoring them &
 \small The sexual behaviors and preferences of individuals are private and are not the business of the authorities\\
 \hline
 \multicolumn{2}{|c|}{Subsidies \small (70 motions)} \\
 \hline
 \small Providing support for [TOPIC] would benefit society, and is therefore a worthwhile use of government money &
 \small There are better ways to make use of public funds\\
 \hline
 \multicolumn{2}{|c|}{Technology$^t$ \small (15 motions)} \\
 \hline
 \small [TOPIC] is better than the older options&
 \small These new technologies are not as reliable as conventional ones\\
 \hline
  \end{tabular}

\begin{tabular}{ |p{3cm}|p{3cm}|  }
 \hline
 \multicolumn{2}{|c|}{Tradition \small (66 motions)} \\
 \hline
 \small A society should respect its traditions and try to avoid changes for the sake of change &
 \small Society should move on with the times, instead of clinging to old and obsolete traditions which are no longer relevant\\
 \hline
 \multicolumn{2}{|c|}{Value of science$^t$ \small (9 motions)} \\
 \hline
 \small Theories which are not based on scientific methods should not be supported &
 \small Not everything can be explained by science\\
 \hline
 \multicolumn{2}{|c|}{Virtual life$^t$ \small (12 motions)} \\
 \hline
 \small The virtual world enriches our lives in ways that other forms do not &
 \small In the virtual world, people lose touch with reality\\
 \hline
 \multicolumn{2}{|c|}{Wealth distribution \small (11 motions)} \\
 \hline
 \small Society has a duty to minimize inequality by allocating resources more evenly &
 \small The way to achieve a fair distribution of wealth is to let it be determined by the market forces\\
 \hline
 \multicolumn{2}{|c|}{Welfare state \small (30 motions)} \\
 \hline
 \small The state has a duty to provide for the social and economic security of its citizens &
 \small State-sponsored welfare is counterproductive and actually exacerbates the problem\\
 \hline
\end{tabular}

\vspace{\baselineskip}
Note the special token [TOPIC] which, during labeling and application, is replaced by the topic of the relevant motion. For example, when labeling the motion (disband, NATO) for the CoPA {\em Framework}, the claims presented to the annotators were {\em NATO works efficiently} and {\em NATO fails to achieve its goals}.

\section{Features engineered for (motion, CoPA) pairs}\label{app:features}
For each CoPA $c$ we manually listed a set $c_m$ of Wikipedia titles as related to it. With this in hand, we define a set of 17 features (listed below) that aim to capture the similarity between the motion and the class. These include similarities between the motion's action and topic and the list of Wikipedia titles as well as similarities between the motion's topic and the topics of other motions in the class (as in {\em KNN} above). In addition to these similarity features, we also included counts-based features. Using this feature a logistic regression classifier was trained, and each CoPA was assigned the score computed by it. 

\subsection{Similarity features}
We associate a motion $m$ with two sets of texts. $m_t = {action, topic}$ is simply the set containing the text of the action and the text of the topic. The second set aims to identify Wikipedia titles related to the topic. Each Wikipedia title linked to in the topic's Wikipedia article is scored by the p-value computed for it for its appearance in the article compared to a set of random articles, using the hypergeometric distribution. $m_w$ is the set of (at most) 10 titles with the lowest p-value.

We also associate each CoPA with two sets of texts. The first is the aforementioned manually-generated list, $c_m$. The second is the set of topics of motions in the CoPA, $c_t$ (when doing leave-one-out analysis, we always ignore occurrences of the topic of the left-out motion).

Given some method to compute similarity between two terms, we define the similarity between two sets of terms as the average over all pairs of terms, one from each set. We employ three types of similarity scores: word2vec (\citet{word2vec}), that of \citet{term-relater}, and cosine similarity of Tf-Idf vectors. All in all this defines 12 similarity features.

In addition, we take all terms in $c_m$ which also appear in the Wikipedia article of the {\em topic}, and take their average Idf score as a 13th similarity feature. 

\subsection{Counts-based features}
For a motion $m = (a,t)$, Let $M_a$ be the set of all motions with action $a$. Let $M_*$ be the set of all motions in our dataset. For $m$ and CoPA $c = (A_c,M_c)$ we define the following four counts-based features:
\begin{enumerate}
    \item $|M_a| / |M_*|$
    \item $|M_a \cap M_c| / |M_a \cup M_c|$
    \item $|M_a \cap M_c| / |M_a|$
    \item $|M_a \cap M_c| / |M_c|$
\end{enumerate}


\end{document}